\documentclass[final,11pt,breaklinks]{colt2021}
\usepackage{times}
\usepackage{multirow}
\usepackage{framed}

\usepackage{amsmath,amssymb,rotating,graphicx,rotating,float}
\usepackage[mathscr]{eucal}

\usepackage{url}

\renewcommand{\rm}{\mathrm}

\newcommand{\X}{\mathcal X}

\newcommand{\T}{\mathcal T}

\newcommand{\nats}{\mathbb{N}}

\newcommand{\reals}{\mathbb{R}}

\newcommand{\E}{\mathbb E}

\DeclareSymbolFont{bbold}{U}{bbold}{m}{n}
\DeclareSymbolFontAlphabet{\mathbbold}{bbold}
\newcommand{\ind}{\mathbbold{1}}

\newcommand{\ProcX}{\mathbb{X}}
\newcommand{\ProcY}{\mathbb{Y}}

\newcommand{\target}{f^{\star}}

\newcommand{\SOKC}{\mathcal{C}_{s}}
\newcommand{\WOKC}{\mathcal{C}_{w}}

\newcommand{\ignore}[1]{}
\newcommand{\private}[1]{}

\newtheorem{problem}{Open Problem}

\title[Online Learning Whenever Online Learning is Possible]{Open Problem: Is There an Online Learning Algorithm\\ That Learns Whenever Online Learning Is Possible?}

\coltauthor{%
\Name{Steve Hanneke} \Email{steve.hanneke@gmail.com} \\ \addr Toyota Technological Institute at Chicago
}

\usepackage{titlesec}
\titlespacing*{\section}{0pt}{-0.5\baselineskip}{\baselineskip}

\begin{document}
\maketitle

\section{Background}
\label{sec:background}

{\vskip -4mm}One of the most classical topics in learning theory is \emph{online learning},
wherein a learning algorithm observes a stream of data points $X_t$ from a space $\X$,
and for each time $t \in \nats$ it makes a prediction $\hat{Y}_t \in \{0,1\}$, after which it observes
the \emph{target} label $Y_t \in \{0,1\}$.  The prediction $\hat{Y}_t$ may depend on
the past observations ($(X_s,Y_s)$, $s < t$) and $X_t$, but nothing else:
that is, $\hat{Y}_t = f_t(X_{1:(t-1)},Y_{1:(t-1)},X_t)$ for a (possibly randomized) function $f_t$.
The objective is to make few \emph{mistakes}:
that is, few times $t$ for which $\hat{Y}_t \neq Y_t$.
In particular, we are most interested in achieving a number of mistakes among $\{(X_t,Y_t)\}_{t \leq T}$ growing \emph{sublinearly} in $T$: that is, $o(T)$ mistakes.
In various versions of this problem,
the sequences $\ProcX := \{X_t\}_{t \in \nats}$ and $\ProcY := \{Y_t\}_{t \in \nats}$
can be either deterministic or random, and either oblivious or adaptive.
Here we focus on the special case where $Y_t = \target(X_t)$ for a fixed unknown \emph{target concept} $\target : \X \to \{0,1\}$,
and where the sequence $\ProcX$ may be random and is independent of any internal randomness in the learning algorithm $f_t$.

Since it is certainly not possible to guarantee few mistakes for \emph{all} pairs $(\ProcX,\target)$,
some restrictions are necessary, and in this respect theories of online learning may be grouped into three categories:
(1) those which allow \emph{arbitrary} sequences $\ProcX$ but restrict the allowed target concepts $\target$ \citep*[e.g.,][]{littlestone:88},
(2) those which restrict both $\ProcX$ and $\target$ \citep*[e.g.,][]{haussler:94},
and (3) those which restrict the sequence $\ProcX$ of points but allow \emph{arbitrary} target concepts $\target$ \citep*[e.g.,][]{stone:77}.
In particular, a classic result in category (1) is that there exists a learning algorithm guaranteeing a \emph{bounded} number
of mistakes for every sequence $\ProcX$ \emph{if and only if} the set of allowed target concepts $\target$ has a
finite \emph{Littlestone dimension} \citep*{littlestone:88} (see \citealp*{ben-david:09}, for the definition of Littlestone dimension).
The works falling in category (2) vary widely in the types of restrictions they impose and the resulting guarantees that are possible.
In particular, note that by introducing restrictions on the (possibly random) sequence $\ProcX$,
we can, to some extent, express the classic theory of \emph{statistical} learning.  For instance,
there exist learning algorithms guaranteeing $O(\log(T))$ mistakes (in expectation) for every
$\ProcX$ that is an \emph{i.i.d.}\ process on $\X$ when
the set of allowed target concepts $\target$ has finite \emph{VC dimension} \citep*{haussler:94}.
Other more-involved restrictions on the pair $(\ProcX,\target)$ have also been considered
\citep*[e.g.,][]{ryabko:06,urner:13,bousquet:21}.

The subject of our present discussion is category (3):
that is, \emph{unrestricted} target concepts $\target$,
but with restrictions on the sequence $\ProcX$.
There has also been significant work in this category.
As a simple example, there exist learning algorithms guaranteeing $o(T)$ mistakes (almost surely) for every
target concept $\target$ and every $\ProcX$ that is an \emph{i.i.d.}\ process on $\X$ \citep*{stone:77,devroye:96,hanneke:21,hanneke:21b}
(see Remark~\ref{rem:measurable} for relevant technical conditions on $\X$).
Indeed, for $\X = \reals^d$, this even holds (in expectation) for the simple $1$-nearest neighbor algorithm \citep*{cover:67,stone:77,devroye:96}.
There is also an extensive literature considering relaxations of the i.i.d.\ assumption,
such as allowing $\ProcX$ to be \emph{stationary ergodic} \citep*{morvai:96,gyorfi:99,gyorfi:02a},
or generally to satisfy a \emph{law of large numbers} \citep*{morvai:99,steinwart:09},
while maintaining this guarantee of $o(T)$ mistakes (either in-expectation or almost surely).

The recent work of \citep*{hanneke:21} unifies and weakens these various restrictions on $\ProcX$,
aiming (in part) to study the \emph{fundamental limits} of this category.
The following definition provides a formal criterion for online learning in this setting.

\begin{definition}
For a (possibly random) sequence $\ProcX = \{X_t\}_{t \in \nats}$,
an online learning algorithm \linebreak$f_t : \X^{t-1} \times \{0,1\}^{t-1} \times \X \to \{0,1\}$ (possibly randomized, independent of $\ProcX$)
is weakly universally consistent under $\ProcX$ if, for every (measurable) $\target : \X \to \{0,1\}$,
\begin{center}
$\E\!\left[ \sum_{t=1}^{T} \ind[ f_t(X_{1:(t-1)},\target(X_{1:(t-1)}),X_t) \neq \target(X_t) ] \right] = o(T)$,
\end{center}
and is strongly universally consistent under $\ProcX$ if, for every (measurable) $\target : \X \to \{0,1\}$,
\begin{center}
$\sum_{t=1}^{T} \ind[ f_t(X_{1:(t-1)},\target(X_{1:(t-1)}),X_t) \neq \target(X_t) ] = o(T) \text{ (a.s.)}$.
\end{center}
\end{definition}

\begin{definition}
For a (possibly random) sequence $\ProcX$, we say (weak/strong) universal online learning is \emph{possible} under $\ProcX$
if there exists an online learning algorithm that is (weakly/strongly) universally consistent under $\ProcX$.
\end{definition}

It is clear that not \emph{every} $\ProcX$ admits universal online learning: for instance,
for $\X = \nats$, universal online learning is not possible under $\ProcX = \{1,2,3,\ldots\}$.
Hence,
to approach the goal of achieving the fundamental limits of online learning in this setting,
\citep*{hanneke:21} introduces a style of reasoning referred to as the \textbf{optimist's decision theory},
described abstractly as follows.

{\vskip 2mm}\paragraph{The Optimist's Decision Theory:} Supposing we are tasked with achieving a given objective $O$ in some scenario,
then already we have implicitly committed to the assumption that achieving objective $O$ is at least \emph{possible} in that scenario:
the \emph{optimist's assumption}.
Since we must commit to this assumption to even begin designing a strategy for achieving objective $O$,
we may \emph{rely} on this assumption in our strategy for achieving the objective.  We are then most interested in
strategies guaranteed to achieve objective $O$ \emph{without any additional assumptions}.  Such a strategy is \emph{universal}
in the most-general sense possible, since the optimist's assumption is \emph{necessary}.  It will achieve the objective $O$
in \emph{all} scenarios where it is possible to do so.	Moreover, such strategies have the satisfying property that, if ever they
fail to achieve the objective, we may rest assured that no other strategy could have succeeded, so that nothing was lost.

{\vskip 2mm}Based on the above reasoning, an online learning algorithm that is universally consistent for every $\ProcX$ that admits universal online learning is called \emph{optimistically universal}:

\begin{definition}
An online learning algorithm is \emph{optimistically (weakly/strongly) universal} if it is \linebreak (weakly/strongly) universally consistent
under every $\ProcX$ such that (weak/strong) universal online learning is possible under $\ProcX$.
\end{definition}

In this context, we present two fundamental questions about online learning,
originally posed in \citep*{hanneke:21}.
The first asks whether there \emph{exists} an optimistically universal online learning algorithm,
while the second asks whether there is a basic property of $\ProcX$ that determines whether universal online learning is possible,
proposing a particular candidate condition for concreteness.
We now turn to these two questions in detail.

\titlespacing*{\section}{0pt}{0.5\baselineskip}{\baselineskip}
\section{Open Problem 1: Optimistically Universal Online Learning}

{\vskip -4mm}The first open problem concerns the existence of optimistically universal online learning algorithms.

\begin{problem}
\label{prob:optimistic-online}
Does there exist an optimistically universal online learning algorithm?~ (in either the weak or strong sense)
\end{problem}

\paragraph{Prize:} I am offering \$5,000 USD for a solution to this problem (be it positive or negative).
If the weak/strong variants are solved in separate works, whichever is solved first will receive the prize.

\begin{remark}[Remark on generality]
\label{rem:measurable}
The general setting considered in \citep*{hanneke:21} allows that $\X$ is any nonempty space equipped with a separable metrizable topology $\T$,
and the measurable sets are specified by the Borel $\sigma$-algebra generated by $\T$.
However, I am willing to award the prizes for any solution general enough to address the case $\X=\reals^d$ with the Euclidean topology.
\end{remark}

\paragraph{Notes:}
The answer to Open Problem~\ref{prob:optimistic-online} is known to be positive
in the special case of \emph{countable} $\X$, or for general $\X$ but with the restriction to \emph{deterministic} sequences $\ProcX$
\citep*{hanneke:21}.  Indeed, as discussed below, in both cases a simple \emph{memorization}-based algorithm suffices.
However, for uncountable $\X$ and general (random) sequences $\ProcX$,
there are simple cases where memorization fails: for instance, $\ProcX$ as any non-atomic i.i.d.\ process.
Thus, the case that remains open concerns \emph{uncountable} $\X$ and general (random) sequences $\ProcX$.
Also note that it is conceivable that
an optimistically strongly universal online learning algorithm is not necessarily also optimistically weakly universal,
since the latter requires universal consistency under a strictly larger family of processes (see below).
Nevertheless, I conjecture that the answers to the weak/strong variants
will be the same.

\section{Open Problem 2: When Is Universal Online Learning Possible?}

{\vskip -4mm}The second open problem concerns characterizing the family of random sequences $\ProcX$
under which universal online learning is possible.  In addition to being intrinsically interesting,
this would likely also be an extremely helpful step toward resolving Open Problem~\ref{prob:optimistic-online}.
To make the problem concrete, \citep*{hanneke:21} proposes the following two conditions.

\begin{definition}
\label{con:okc}
\newline $\bullet$ Let $\WOKC$ denote the family of all (possibly random) sequences $\ProcX = \{X_t\}_{t \in \nats}$ such that
every disjoint sequence $\{A_i\}_{i \in \nats}$ of measurable sets satisfies
$\E\!\left[ \left| \{ i \in \nats : X_{1:T} \cap A_i \neq \emptyset \} \right| \right] = o(T)$.\\
\\$\bullet$ Let $\SOKC$ denote the family of all (possibly random) sequences $\ProcX = \{X_t\}_{t \in \nats}$ such that
every disjoint sequence $\{A_i\}_{i \in \nats}$ of measurable sets satisfies
$\left| \{ i \in \nats : X_{1:T} \cap A_i \neq \emptyset \} \right| = o(T) \text{ (a.s.)}$.
\end{definition}

\noindent We then have the following open problem.

\begin{problem}
\label{prob:okc-equivalence}
\newline$\bullet$~ Is $\WOKC$ equal to the set of all $\ProcX$ such that weak universal online learning is possible under $\ProcX$?
\\$\bullet$~ Is $\SOKC$ equal to the set of all $\ProcX$ such that strong universal online learning is possible under $\ProcX$?
\end{problem}

\paragraph{Prize:} I am offering \$1,000 USD for a solution to either of these questions (be it positive or negative).
If the weak/strong variants are solved in separate works, whichever is solved first will receive the prize.
Additionally, I note that Remark~\ref{rem:measurable} also applies to this problem.

\paragraph{Notes:}
The work \citep*{hanneke:21} establishes that $\ProcX \in \WOKC$ or $\ProcX \in \SOKC$ are \emph{necessary} for
weak or strong, respectively, universal online learning to be possible under $\ProcX$.	Moreover, in the special case
of countable $\X$, or for general $\X$ but with the restriction to deterministic sequences $\ProcX$, \citep*{hanneke:21} also shows that
$\ProcX \in \WOKC$ or $\ProcX \in \SOKC$ are \emph{sufficient} for weak or strong, respectively, universal online learning
to be possible under $\ProcX$.
That is, both questions in Open Problem~\ref{prob:okc-equivalence} have \emph{positive} answers for countable $\X$, or for general $\X$ with the restriction to deterministic $\ProcX$.
Indeed, it is an easy exercise to verify that the simple \emph{memorization} algorithm is weakly or strongly universally consistent in these cases, when $\ProcX \in \WOKC$ or $\ProcX \in \SOKC$, respectively.
Thus, the case that remains open in Open Problem~\ref{prob:okc-equivalence} concerns
whether $\ProcX \in \WOKC$ or $\ProcX \in \SOKC$ are sufficient conditions for weak or strong, respectively, universal online learning
to be possible under $\ProcX$ for \emph{uncountable} $\X$ and general (random) sequences $\ProcX$.
The route to proving such a result (positively) would be to construct an online learning algorithm and show
that it is weakly or strongly universally consistent under every $\ProcX$ in $\WOKC$ or $\SOKC$, respectively.
Note that, unlike Open Problem~\ref{prob:optimistic-online}, an algorithm sufficient to positively resolve Open Problem~\ref{prob:okc-equivalence}
may even depend on the distribution of $\ProcX$.

\citep*{hanneke:21} also discusses relations between the sets involved in Open Problem~\ref{prob:okc-equivalence}.
It is clear that $\SOKC \subseteq \WOKC$.
Also, for any $\ProcX$, if strong universal online learning is possible then weak universal online learning must also be possible.
However, supposing $\X$ is infinite, \citep*{hanneke:21} gives an example $\ProcX$
in $\WOKC \setminus \SOKC$, so that the two sets are not equivalent.
Moreover, \citep*{hanneke:21} shows that
weak universal online learning \emph{is} possible under this $\ProcX$,
but \emph{strong} universal online learning \emph{is not} possible
under this $\ProcX$.  Thus, the sets of (random) sequences $\ProcX$
under which universal online learning is possible in the weak and strong senses
are \emph{not equal}.

\section{Connections to Related Settings}

{\vskip -4mm}The work of \citep*{hanneke:21} considers three learning settings: inductive, self-adaptive, and online.
The inductive setting is most-familiar to the statistical learning literature, where a learning
algorithm observes a finite \emph{training} set and then produces a fixed hypothesis that is then
used for all future predictions.
The self-adaptive setting differs only in that it allows the learner to update its hypothesis based on
the \emph{unlabeled} data it has made predictions on so far.
The interested reader is referred to that work for the precise definitions.
\citep*{hanneke:21} proves that there \emph{do} exist optimistically universal \emph{self-adaptive}
learning algorithms, meaning that they are universally consistent for all $\ProcX$ such that universal \emph{self-adaptive} learning is possible under $\ProcX$ (both weak and strong).
On the other hand, \citep*{hanneke:21} also proves that optimistically universal \emph{inductive} learning is \emph{impossible} (both weak and strong).
Moreover, \citep*{hanneke:21} provides a concise characterization of the family of all (possibly random) sequences $\ProcX$
such that (weak/strong) universal (inductive/self-adaptive) learning is possible under $\ProcX$.

That work also makes connections between self-adaptive learning and online learning,
providing a technique to convert any self-adaptive learning algorithm into an online learning algorithm,
while preserving consistency.  In particular,
applying this conversion to the optimistically universal self-adaptive learning
algorithm provides
an online learning algorithm that is strongly universally consistent
under \emph{every}
$\ProcX$ under which universal self-adaptive learning is possible.
However, \citep*{hanneke:21} also shows that for any infinite $\X$, there exist sequences $\ProcX$ (even deterministic)
for which universal online learning is possible but universal self-adaptive learning is not possible.
Thus, new techniques are needed to understand the sufficient conditions for universal online learning (Open Problem~\ref{prob:okc-equivalence}),
and to approach the question of optimistically universal online learning (Open Problem~\ref{prob:optimistic-online}).

\bibliography{learning}

\end{document}